\documentclass[runningheads]{llncs}
\usepackage{graphicx}
\usepackage[symbol]{footmisc}

\usepackage{float}
\usepackage{mathtools}
\usepackage{amssymb, amsmath, latexsym}
\usepackage{nicefrac}
\usepackage{hhline}
\usepackage{makecell}
\usepackage{multirow}
\usepackage{wrapfig}

\usepackage{colortbl}
\definecolor{bgcolor}{rgb}{0.8,1,1}
\definecolor{bgcolor2}{rgb}{0.68,0.78,0.81}
\usepackage{threeparttable}

\usepackage{pifont}
\usepackage[colorinlistoftodos,bordercolor=orange,backgroundcolor=orange!20,linecolor=orange,textsize=scriptsize]{todonotes}
\usepackage{algorithm}\usepackage{algpseudocode}

\allowdisplaybreaks
\graphicspath{ {images/} }
\usepackage[justification=centering]{caption}
\usepackage{subcaption}

\usepackage{tabularx} 
\usepackage{xcolor} 

\definecolor{LightGray}{gray}{0.9}

\begin{document} 
\title{Activations and Gradients Compression for Model-Parallel Training\thanks{The research of A. Beznosikov was supported by Russian Science Foundation (project No. 23-11-00229).}}
\titlerunning{Activations and Gradients Compression for Model-Parallel Training}
%

\author{Mikhail Rudakov\inst{1, 2} \and
Aleksandr Beznosikov\inst{1,2} \and
Yaroslav Kholodov\inst{1} \and
Alexander Gasnikov\inst{1,2}
}
\authorrunning{M. Rudakov, A. Beznosikov, Y. Kholodov, A. Gasnikov}
%
\institute{Innopolis University, Innopolis, Russia, \and
Moscow Institute of Physics and Technology, Moscow, Russia
}
\maketitle              
\begin{abstract}

Large neural networks require enormous computational clusters of machines. Model-parallel training, when the model architecture is partitioned sequentially between workers, is a popular approach for training modern models. Information compression can be applied to decrease workers communication time, as it is often a bottleneck in such systems. This work explores how simultaneous compression of activations and gradients in model-parallel distributed training setup affects convergence. We analyze compression methods such as quantization and TopK compression, and also experiment with error compensation techniques. Moreover, we employ TopK with AQ-SGD per-batch error feedback approach. We conduct experiments on image classification and language model fine-tuning tasks.
Our findings demonstrate that gradients require milder compression rates than activations. We observe that $K=10\%$ is the lowest TopK compression level, which does not harm model convergence severely. Experiments also show that models trained with TopK perform well only when compression is also applied during inference. We find that error feedback techniques do not improve model-parallel training compared to plain compression, but allow model inference without compression with almost no quality drop. Finally, when applied with the AQ-SGD approach, TopK stronger than with $ K=30\%$ worsens model performance significantly.


\end{abstract}

\section{Introduction}
\label{sec:introduction}

Neural networks have recently emerged as an essential field of computer science research. Hundreds of articles are published yearly on machine learning, including computer vision, natural language processing, reinforcement learning, and generative models.
One of the reasons for such success is the tremendous growth of the model and dataset sizes. Recent examples include GPT-4~\cite{openai2023gpt4} and BLOOM~\cite{scao2022bloom}, with up to billions of trainable parameters. Such models also require terabytes of data for training. Recent ROOTS text corpus \cite{ROOTSlaurenccon2022bigscience}, used for BLOOM training, is 1.6 TB large. Therefore, such models require a lot of CPU and GPU memory. Such computational resources are available only to a few large institutions and companies that can afford big servers with large network bandwidth. Nevertheless, smaller companies and research groups can also attempt to use modern architectures by pooling computational resources with each other over the Internet.

In any case, training large models is impossible without using several computational devices, and hence without using a distributed learning approach~\cite{verbraeken2020survey}. In distributed machine learning, data-parallel and model-parallel are two major parallelization paradigms that allow neural networks to be trained on several machines. In data parallelism, the training data is split between worker computers~\cite{krizhevsky2014one}. Each worker contains a full copy of the model and updates it during each training step according to averaged change from all the workers. Shallue et al. in the work \cite{shallue2018measuring} show that data parallelism does not negatively affect model quality with nearly the same number of training iterations. Data-parallel model training is widely used in large models training frameworks: Horovod \cite{sergeev2018horovod}, Pytorch Data-Parallel \cite{li2020pytorch}.

The second approach, model parallelism, or pipeline parallelism (MP/PP), is orthogonal. In model parallelism, the model itself is divided into blocks of layers, each stored on a separate machine. During training, workers are organized into a chain, where adjacent nodes communicate the layers activations in the forward pass and gradients in the backward pass. The model-parallel technique has been extensively used to train large models in systems like Megatron \cite{shoeybi2019megatron}, Deepspeed \cite{rasley2020deepspeed}, Petals \cite{borzunov2022petals}. Moreover, the model-parallel approach usually uses the pipelining technique. It allows distributed systems to use resources efficiently by splitting each batch into micro-batches, which are then pipelined on the block-divided model as in, for example, Gpipe \cite{huang2019gpipe}, Xpipe \cite{guan2019xpipe}, PipeDream \cite{harlap2018pipedream}.

However, slow communication becomes a problem in model parallelism \cite{diskin2021distributed}. During a forward pass, workers send vectors of layers activations. During a backward pass, vectors of respective gradients are sent to update model parameters. Communication time may be a bottleneck, leaving workers idle waiting for the data, especially when model-parallel training happens on machines located worldwide over slow network connection.

Research in the area of compression in model-parallel setup is present in the literature. Various quantization schemes are applied to activations~\cite{fu2020dontwastebits,evans2021acgc,dettmers2022llm} or gradients~\cite{song2023optimus}. Sparsification techniques, which are popular in data-parallel case for gradients compression~\cite{stich2018sparsifiedTopkRandk,beznosikov2020biased}, have only a few results in the model-parallel case for  activations compression~\cite{bian2023doesactivationcompression,gupta2021trainingrecommender}. Several works also utilize error compensation techniques in the model-parallel setting for activations~\cite{wang2022fineAQSGD} and gradients~\cite{song2023optimus} for efficient convergence.
Despite the work done on model-parallel compression, results tend to be independent and mostly empirical, obtained for activations or gradients separately.

In this work, we present experiments on model-parallel compression. Our focus is to study simultaneous compression of activations and gradients for various compression operators. Moreover, we experiment with existing error feedback approaches to test their applicability with sparsification compression.

The work is organized as follows. Section~\ref{section:methodology} describes experiment design and compression operators and techniques used for experiments with model parallelism. Section~\ref{section:results} presents and discusses results of model-parallel experiments in various compression scenarios. Section~\ref{section:literature_review} reviews related works on activations and gradients compression in distributed learning. Finally, Section~\ref{section:conclusion} concludes the work.

\subsection{Contributions}

\hspace{0.35cm} \textbf{Simultaneous activation and gradient compression:} We conduct experiments on compressing both activations and gradients in model-parallel training and empirically evaluate convergence of models trained with quantization and TopK compressors, showing that:
    \begin{enumerate}
        \item In the case of quantization, gradients are more sensitive to compression than activations, requiring less compression for good convergence;
        \item The strongest TopK compression level that does not harm the convergence is Top10\%, and compression has to be applied during inference to produce comparable validation quality.
    \end{enumerate}

\textbf{Experiments with error feedback:} We apply various error compensation techniques to activations and gradients compression in the model-parallel setup, including EF and EF21, and show that:
    \begin{enumerate}
        \item Models trained with TopK compression and error feedback produce good validation results even with no compression applied;
        \item When applied to both activations and gradients, error feedback techniques do not improve model convergence.
    \end{enumerate}

\textbf{AQ-SGD with TopK compression:} We evaluate how AQ-SGD~\cite{wang2022fineAQSGD} error feedback for activations works with TopK compression and conclude it does not improve model convergence compared to plain TopK compression.


\section{Methodology}
\label{section:methodology}

In this section, we describe the overall design of model-parallel compression experiments and characteristics of evaluated compression operators.

\subsection{Experiment Design: Model-Parallel Training with Compression}
\label{section:methodology_design}

\begin{figure}[h!]
         \centering
         \includegraphics[width=0.7\linewidth]{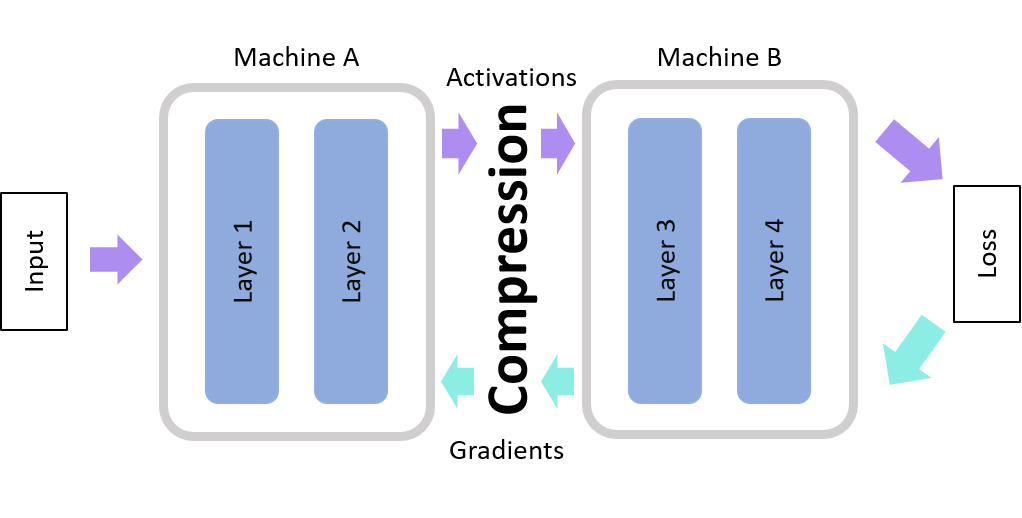}
         \caption[Model parallel training example.]{Model parallel training example. Model parallelism degree is two, with one compression block. Activations are compressed in the forward pass, and gradients are compressed in the backward pass.}
         \label{fig:methodology-design}
 \end{figure}

In our experiments, we conduct training and fine-tuning of large neural networks using model parallelism approach with compression between pipeline stages. Figure~\ref{fig:methodology-design} presents this approach, where the whole model is partitioned on two devices with single communication stage in between (model parallelism degree is two). Adjacent workers in the pipeline exchange activations in the forward pass and gradients of these activations in the backward pass.

We use compression operators to send less data, which reduces potential communication overhead. In our experiments, we compress both activations and gradients to evaluate the limits at which compression can be used without compromising model convergence and quality.

Since our work focuses only on these two objectives, we do not conduct real model-parallel training on several machines. Instead, we integrate compression to the model directly, compressing activations and gradients between pipeline blocks, where communication happens in a real distributed system. This approach is equivalent to model-parallel training in terms of convergence analysis.

To reduce communication overhead, we apply various compression operators to activations and their respective gradients. We analyze convergence of quantization and TopK compression operators, and also use error feedback and AQ-SGD techniques with TopK compression.

\subsection{Quantization}
\label{subsec:methodology_quantization}
Quantization is a compression technique that involves mapping floating point numbers to a finite set of integer values. In our experiments, we utilize uniform $k$-bit quantization with scaling. Input vector with floating values is mapped to $[0;1]$ interval with min-max scaling and then quantized to $k$ levels. Decompression transforms these values back to original scale.
In our experiments, we evaluate quantization to 2, 4, 6, 8 bits for activations and gradients.

\subsection{TopK compression}
\label{subsec:methodology_topk}
TopK compression is a sparsification method with proven convergence rates for data-parallel setup \cite{beznosikov2020biased}.
The TopK operator selects the largest K\% values (in absolute value) from the input vector, setting other values to zero. 
This compression technique selects only the most critical data to send, which are supposed to be the largest activations or gradients. 

We evaluate several levels of TopK compression (Top50\%, Top30\%, Top20\%, Top10\%, Top5\%) for each task, compressing both activations and gradients.

\subsection{Error Feedback}
\label{subsec:methodology_ef}

Error feedback (EF) is an error compensation technique widely used in data-parallel training with compression applied \cite{seide20141,mishchenko2019distributedDIANA,richtarik2021ef21}. EF technique compensates errors introduced by compression by sending the accumulated error during next communication rounds. Such approach ensures that eventually all information contributes to the learning process.

The original version of EF, proposed by Seide et al. \cite{seide20141}, saves compression error and adds it during the next communication round. Given vector $X$ and compression operation $\mathcal{C}$, error is computed as $e = X - \mathcal{C}(X)$. During the next iteration, the compressed message is the input value added to the previous error, i.e., $\mathcal{C}(X + e)$. Error is propagated to update model parameters on each iteration.

An advanced version of error feedback, EF21, was introduced by Richtarik et al. \cite{richtarik2021ef21}. It features the compression of the current and previous parameter states' differences. As the parameter values tend to converge, the difference between two consecutive iterations' parameters converge to zero, and thus compressing such values would produce less communication error. Given activations vector $X_{i+1}$ and previously sent value $g_i$, the communicated message is $\mathcal{C}(X_{i+1}-g_i)$, and buffer value is updated as $g_{i+1} = g_i + \mathcal{C}(X_{i+1}-g_i)$. 

Based on ideas of error feedback techniques, we also introduce EF-mixed compression schema for TopK compression. The idea is to compress the largest $K/2$\% values from the input vector $X$ and $K/2$\% of the current error buffer. After that, the input vector and buffer are added and sent, resulting in the same $K$\% non-zero values. This approach tries to get the largest and most important input values and the largest communication errors for activations and gradients.

In the experiments, we use global error buffer, meaning the accumulated error is added to the next batch.

\subsection{AQ-SGD compression}
\label{subsec:aq-sgd}
AQ-SGD \cite{wang2022fineAQSGD} is a technique designed to work with model parallelism using a variant of error feedback to compensate compression errors. Similarly to EF21, AQ-SGD compresses and communicates the change of activation values throughout the training process. However, the error buffer is individual for each batch in the training set, resulting in large memory footprint. The original work uses quantization for forward and backward compression, with the AQ-SGD technique applied only for activations.

We use the AQ-SGD approach combined with TopK compression in our experiments. We evaluated whether AQ-SGD remains effective under biased TopK compression applied for activation and gradient, instead of the quantization in the original work. We test Top50\%, Top30\%, Top20\%, and Top10\% compression with AQ-SGD error feedback for activations and plain TopK compression for gradients.


\section{Results}
\label{section:results}

In this section, we present and evaluate the results of our experiments with various activation compression techniques in model-parallel neural network training. We present experiments with the ResNet-18 network and CIFAR-10 dataset. These experiments include quantization and TopK compression, usage of error feedback (EF) techniques, and AQ-SGD experiments. We also discuss experiments on fine-tuning GPT-2 on the Wikitext dataset with TopK compression. Our experiment code is available at GitHub\footnote{\url{https://github.com/Glemhel/ActivationsGradientsCompressionForMP}}.

\subsection{Training ResNet18 on CIFAR-10}
\label{sec:resnet-cifar}
The first series of experiments involved training ResNet18~\cite{he2016deepresnset} convolutional neural network on CIFAR-10~\cite{krizhevsky2009cifar} dataset of labeled images. Image classification quality is measured with training accuracy; also train and test loss are reported.

Experiments were conducted on a single Tesla P100-PCIE-16GB GPU with model parallelism degree of 4.
Training epochs was set to 100 epoch with a batch size of 100. SGD optimizer with momentum 0.9 and weight decay 5e-4 was used. The learning rate is controlled by a cosine annealing scheduler, with an initial value of 0.01 and $T_{max} = 200$.

Our implementation is based on the repository by kuangliu\footnote{\url{https://github.com/kuangliu/pytorch-cifar}}, which compares various image classification models on the CIFAR-10 dataset.

As a baseline, training with no compression was conducted. It achieved 93.0\% average test accuracy over five runs.

\begin{table}[h]
    \caption[Quantization Experiments Training Results on ResNet18 and CIFAR-10]
    {Quantization Experiments Training Results\\on ResNet18 and CIFAR-10}
    \label{table:results-resnet-quantization}
    \begin{center}
    \begin{tabularx}{0.75\textwidth}{X c c}
        \hline
        \makecell[l]{Compression\\Mode} & \makecell[l]{Test accuracy (\%),\\compression off} & \makecell[l]{Test accuracy (\%),\\with compression}\\
        \hline
        No compression & \textbf{93.00}  & \textbf{93.00} \\
        fw4-bw8 & \textbf{93.10}  & \textbf{92.95} \\
        fw4-bw6 & \textbf{91.93}  & \textbf{91.76} \\
        fw4-bw4 & 83.39  & 82.66 \\
        fw4-bw2 & 65.11  & 64.27 \\
        fw2-bw8 & 77.09  & \textbf{92.05} \\
        fw2-bw6 & 84.87  & \textbf{91.59} \\
        fw2-bw4 & 81.32  & 83.92 \\
          \hline
    \end{tabularx}
    \end{center}
    \begin{center}
    \small
    fw[A]-bw[B] means compressing activations to A bits, gradients to B bits.\\Each cell represents best result over 5 runs, each run for 100 epochs.\\ Model-parallel degree is 4, with 3 compression operations used.
    \end{center}
\end{table}

\begin{figure}[!htbp]
\captionsetup[subfigure]{justification=centering}
     \begin{subfigure}{0.33\textwidth}
         \centering
         \includegraphics[width=\linewidth]{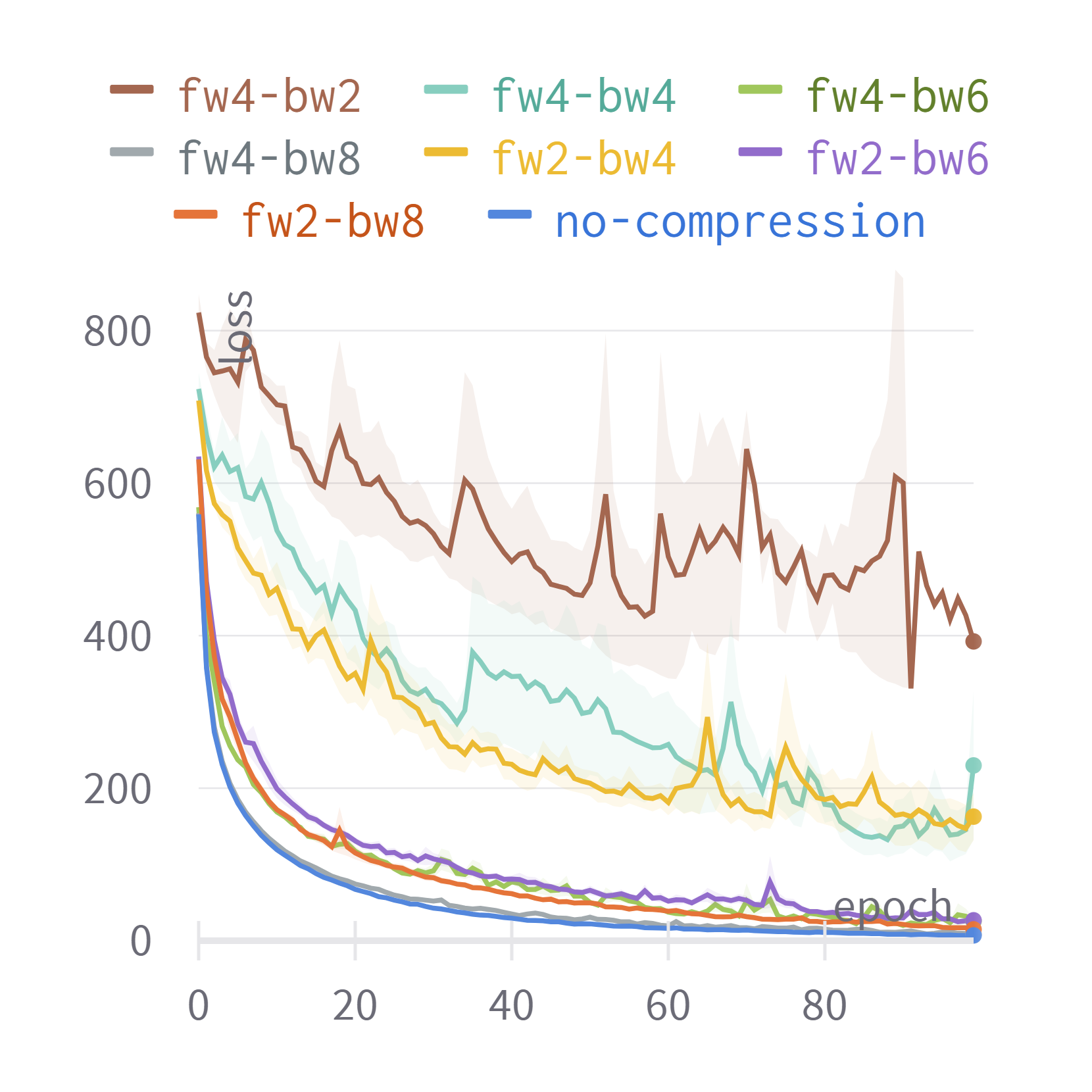}
         \caption{Train loss\\\phantom{..}}
         \label{fig:results-resnet-quantization-trainloss}
     \end{subfigure}%
     \hfill
     \begin{subfigure}{0.33\textwidth}
         \centering
         \includegraphics[width=\linewidth]{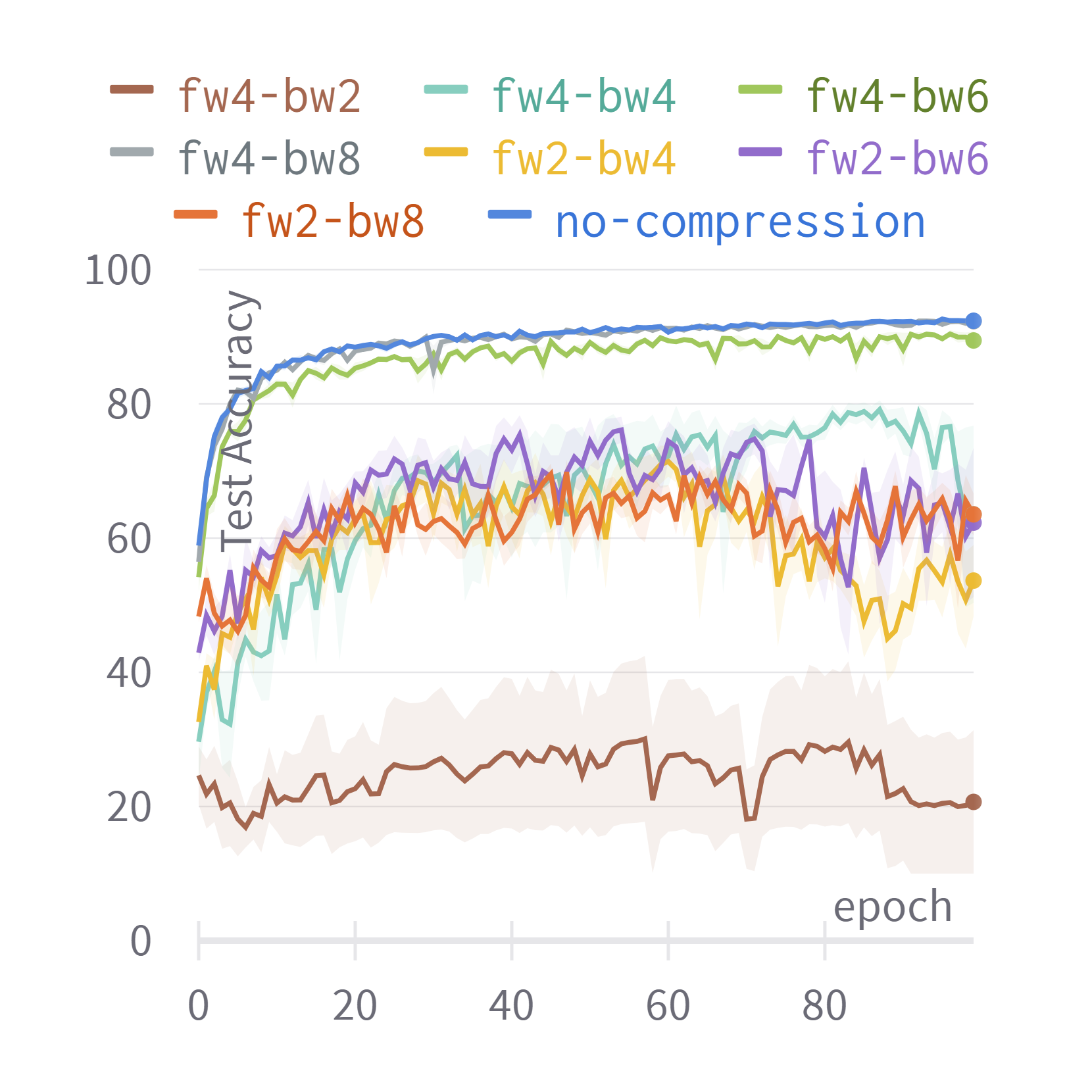}
         \caption{Test accuracy, \\compression off}
         \label{fig:resulsts-resnet-quantization-testacc-nocompression}
     \end{subfigure}%
     \hfill
     \begin{subfigure}{0.33\textwidth}
         \centering
         \includegraphics[width=\linewidth]{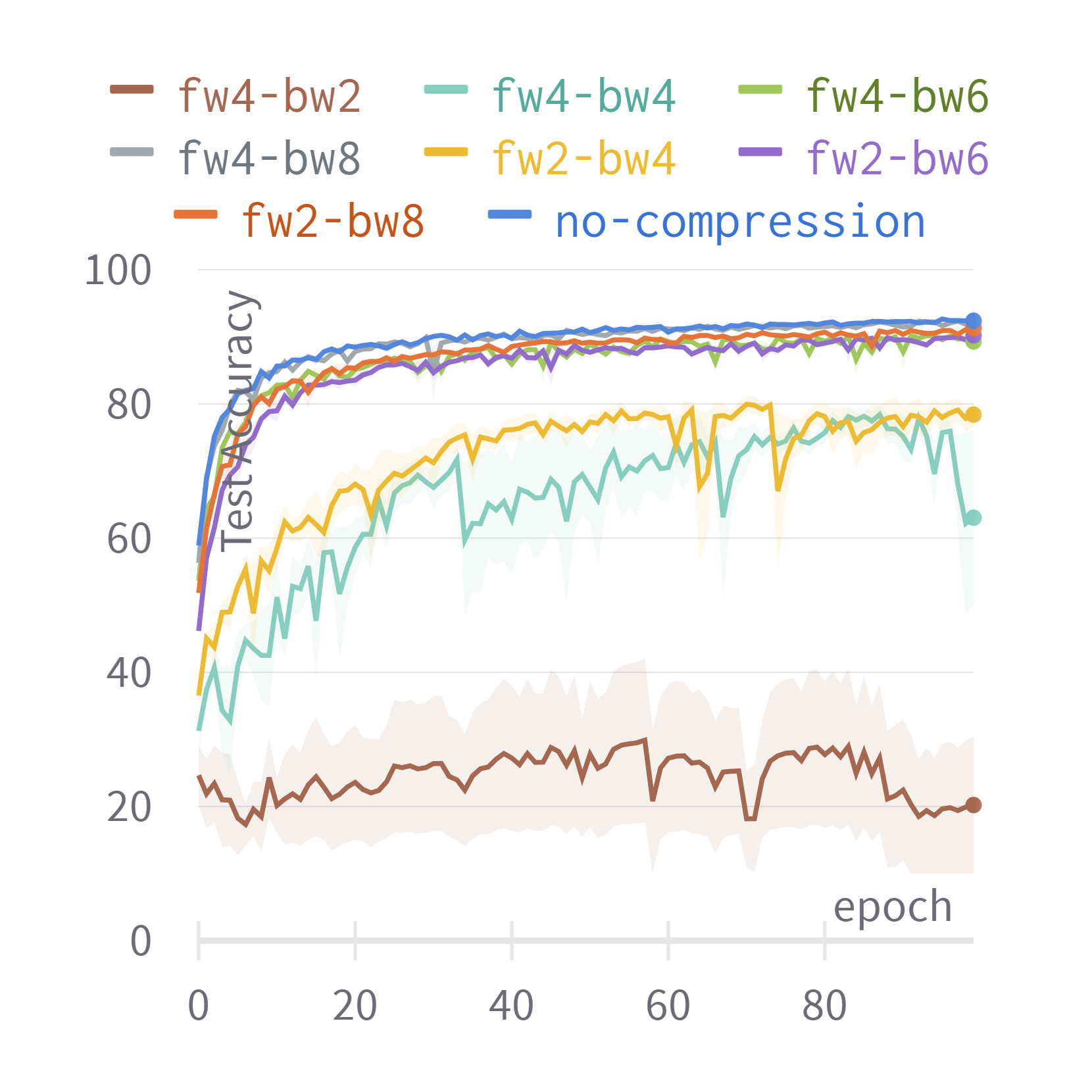}
         \caption{Test accuracy, \\with compression}
         \label{fig:results-resnet-quantization-testacc-withcompression}
     \end{subfigure}%
    \caption[Quantization experiments convergence graphs on ResNet18 and CIFAR-10.]
    {Quantization experiments convergence on ResNet18 and CIFAR-10.\\fw[A]-bw[B] means compressing activations to A bits, gradients to B bits.\\Each line is average $\pm$ standard error over 5 runs.}
    \label{fig:results-resnet-quantization}
\end{figure}

\subsubsection{Quantization}
\label{subsec:resnet-quantization}
Figure~\ref{fig:results-resnet-quantization} shows train loss and uncompressed and compressed accuracy on the test set for quantization experiments. Table~\ref{table:results-resnet-quantization} presents the best test accuracy achieved for each compression experiment.

Firstly, we observe that gradients are more sensitive to quantization than activations. Test accuracy comparable to uncompressed scenario is achieved only when gradients are quantized to at least 6 bits, while activations may be quantized to 2 or 4 bits. Such a result aligns with the results described in \cite{wang2022fineAQSGD,bian2023doesactivationcompression}, where the weak performance of 2 and 4 bits gradients compression is reported. The distinction between compressing activations and gradients may lie in their impact. Compressing gradients can seriously disrupt optimization and convergence, while compressing activations has a milder effect, as long as the model can adapt to the compressed distribution.

Secondly, we report a notable difference in test accuracy with and without compression. Model test accuracy for the compressed case is 7-15 percentage points higher than inference without compression for quantization with activations compressed to 2 bits. We conclude that compression becomes part of the model because not using compression decreases model performance. When uncompressed activations are passed into the model, they unpredictably change the model's output.

\begin{table}[!h]
    \caption[TopK Experiments Training Results on ResNet18 and CIFAR-10]
    {TopK Experiments Training Results \\on ResNet18 and CIFAR-10}
    \label{table:results-resnet-topK}
    \begin{center}
    \begin{tabularx}{0.75\textwidth}{X c c}
        \hline
        \makecell[l]{Compression\\Mode} & \makecell[l]{Test accuracy (\%),\\compression off} & \makecell[l]{Test accuracy (\%),\\with compression}\\
         \hline
        No compression & \textbf{93.00}  & \textbf{93.00} \\
         Top 50\% & \textbf{93.16}   & \textbf{93.38}     \\
         Top 30\% & \textbf{91.26}    & \textbf{93.06}    \\
         Top 20\% & 86.74    & \textbf{92.73}    \\
         Top 10\% &  75.89      & \textbf{91.87}   \\
         Top 5\% & 55.09       & 90.01    \\
         Top 2\% & 49.95       & 85.52    \\
          \hline
    \end{tabularx}
    \end{center}
    \begin{center}
    \small
    Each cell represents best result over 5 runs, each run for 100 epochs.\\ Model-parallel degree is 4, with 3 compression operations used.\\Activations and gradients are compressed independently.
    \end{center}
\end{table}

\begin{figure}[!h]
\captionsetup[subfigure]{justification=centering}
     \begin{subfigure}{0.33\textwidth}
         \centering
         \includegraphics[width=\linewidth]{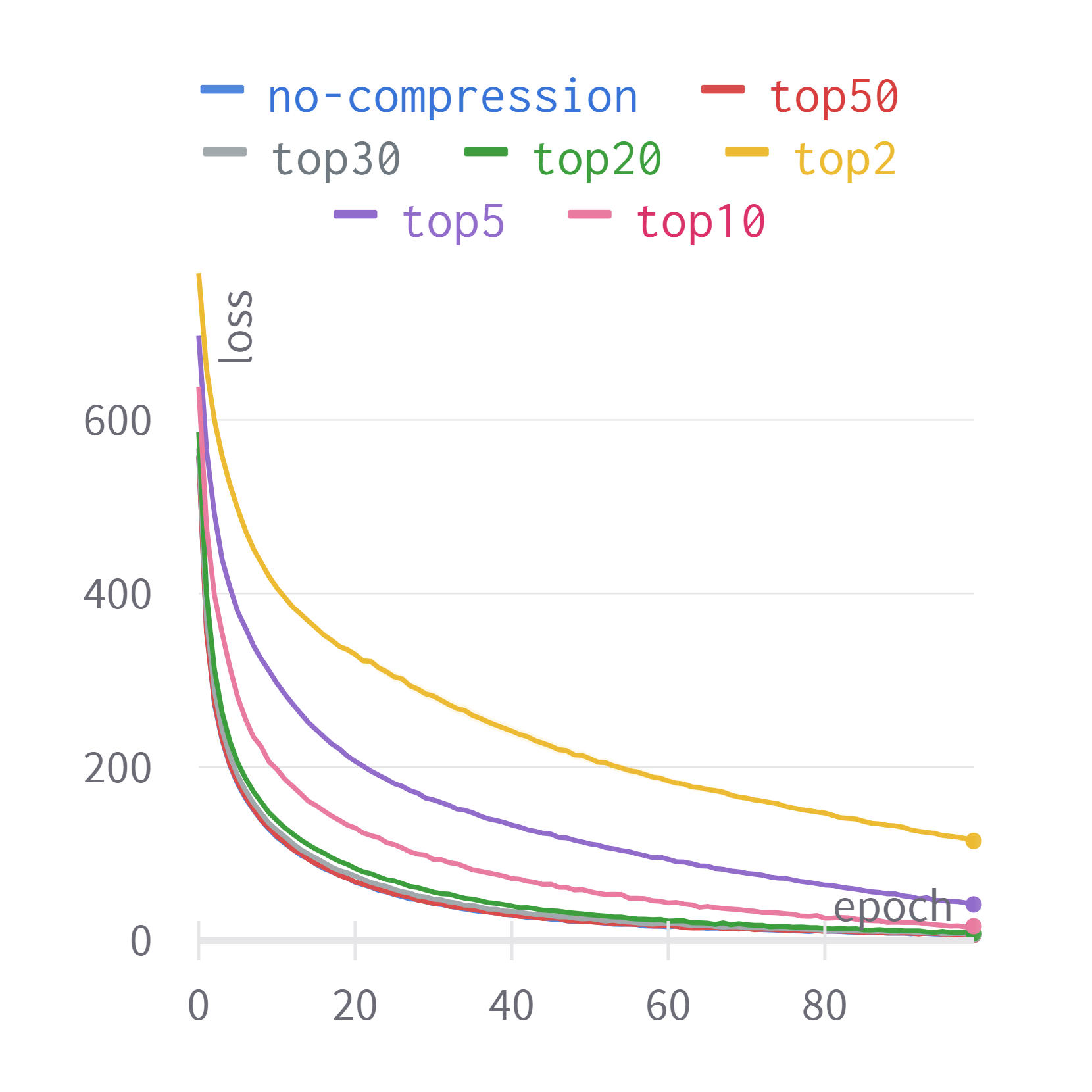}
         \caption{Train loss\\\phantom{..}}
         \label{fig:results-resnet-topk-trainloss}
     \end{subfigure}%
     \hfill
     \begin{subfigure}{0.33\textwidth}
         \centering
         \includegraphics[width=\linewidth]{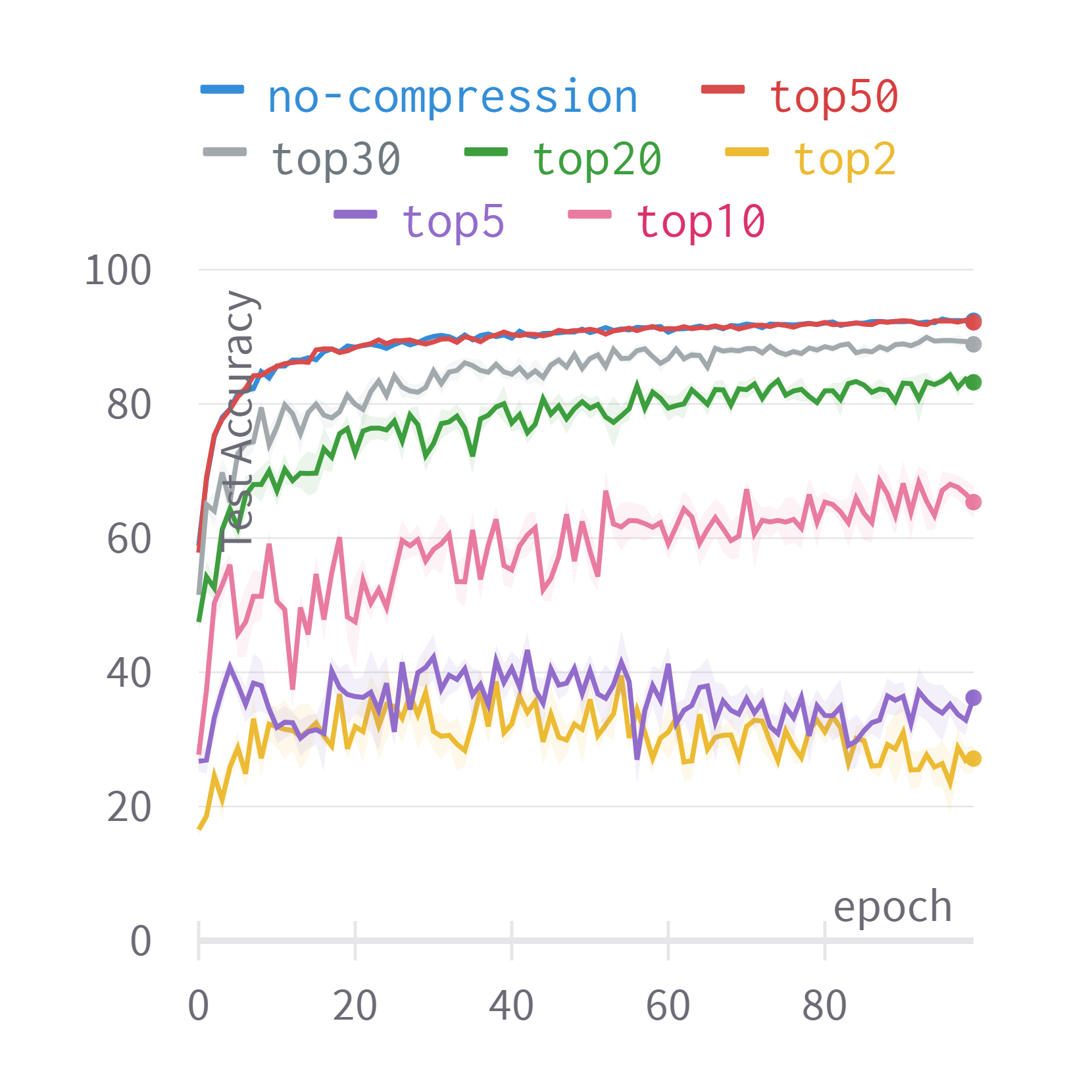}
         \caption{Test accuracy, \\compression off}
         \label{fig:results-resnet-topk-testacc-nocompression}
     \end{subfigure}%
     \hfill
     \begin{subfigure}{0.33\textwidth}
         \centering
         \includegraphics[width=\linewidth]{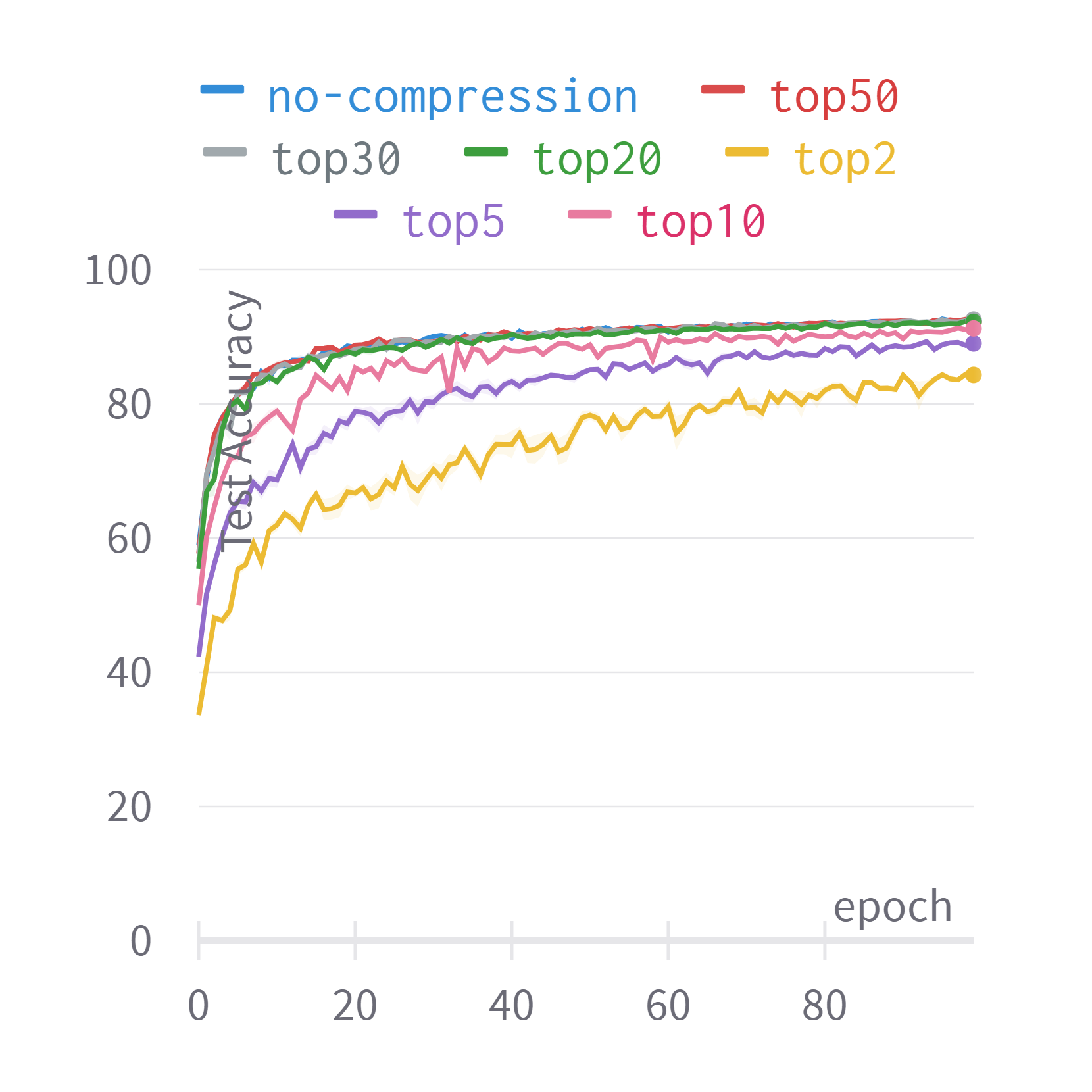}
         \caption{Test accuracy, \\with compression}
         \label{fig:results-resnet-topk-testacc-withcompression}
     \end{subfigure}%
    \caption[TopK experiments on ResNet18 and CIFAR-10.]
    {TopK experiments convergence on ResNet18 and CIFAR-10.\\Each line is average $\pm$ standard error over 5 runs. Model-parallel degree is 4, with 3 compression operations used. Activations and gradients are compressed independently.}
    \label{fig:results-resnet-topk}
\end{figure}

\subsubsection{TopK Compression}
\label{subsec:resnet-topk}
Results of experiments with TopK compression are provided in Table~\ref{table:results-resnet-topK}, with learning curves displayed in Figure~\ref{fig:results-resnet-topk}.
Similarly to the quantization case, we observe that inference with compression results in better test accuracy than without compression for TopK compression. While compressed inference test accuracy stays satisfactory up to Top10\% compression, uncompressed test accuracy is below 90\% already for Top20\%. In the TopK compression case, activations that are usually set to 0 during compression change the model behavior significantly when no compression is applied. Therefore, for a model trained with compression, the application of compression during inference is obligatory to achieve good results. To the best of our knowledge, no previous works considered such aspect of model-parallel communication compression. Moreover, our results demonstrate that Top10\% compression is a good technique for application in convolutional neural networks such as ResNet.

\begin{table}[!h]
    \caption[Experiments with Error Feedback Training Results on ResNet18 and CIFAR-10]
    {Error Feedback Experiments Training Results \\on ResNet18 and CIFAR-10}
    \label{table:results-ef}
    \begin{center}
    \begin{tabularx}{0.85\textwidth}{X c c}
        \hline
        \makecell[l]{Compression\\Mode} & \makecell[l]{Test accuracy(\%), \\compression off} & \makecell[l]{Test accuracy(\%), \\with compression}\\
         \hline
         No compression & \textbf{93.00}  & \textbf{93.00} \\
         \makecell[l]{EF + Top 10\%, warmup 20} & \textbf{91.02}   & 89.85     \\
         \makecell[l]{EFmixed + Top 10\%, warmup 20} & 61.37   & 90.97     \\
         EF21 + Top 5\%& 89.37    & 89.29     \\
         EF21 + Top 10\% & 90.83    & \textbf{91.19}    \\
         \makecell[l]{EF21 + Top 10\%, warmup 20} &  90.71  & \textbf{91.77}    \\
         \hline
    \end{tabularx}
    \end{center}
    \begin{center}
    \small
    Each cell is a single run for 100 epochs.\\Warmup 20 runs use uncompressed baseline weights after 20 epochs.\\Model-parallel degree is 4, with 3 compression operations used.\\Activations and gradients are compressed independently, with global EF batch\\ buffer for each compression operator.
    \end{center}
\end{table}

\subsubsection{Compression with Error Feedback}
\label{subsec:discussion_EF}
\begin{figure}[t]
\captionsetup[subfigure]{justification=centering}
     \begin{subfigure}{0.33\textwidth}
         \centering
         \includegraphics[width=\linewidth]{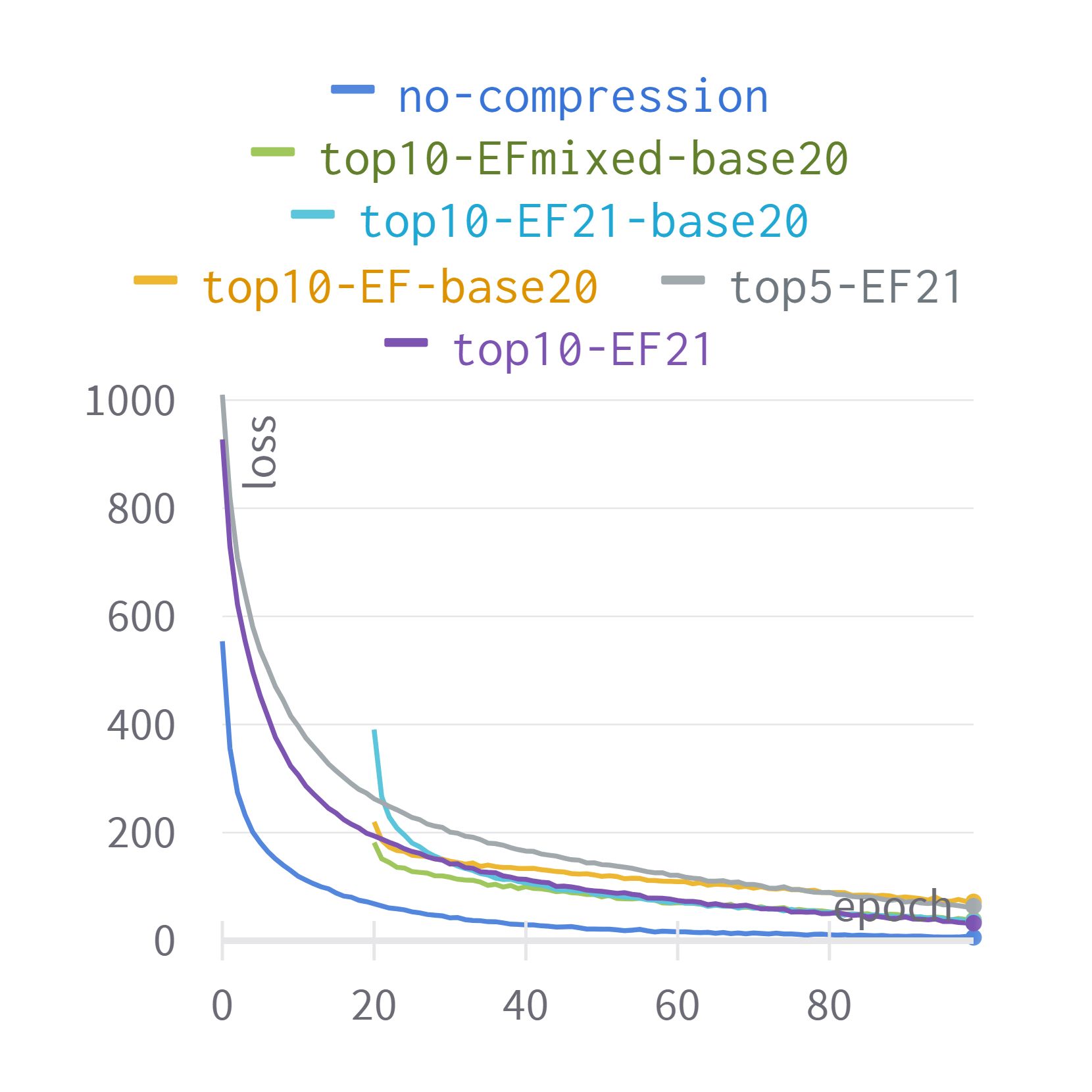}
         \caption{Train loss\\\phantom{..}}
         \label{fig:ef-trainloss}
     \end{subfigure}%
     \hfill
     \begin{subfigure}{0.33\textwidth}
         \centering
         \includegraphics[width=\linewidth]{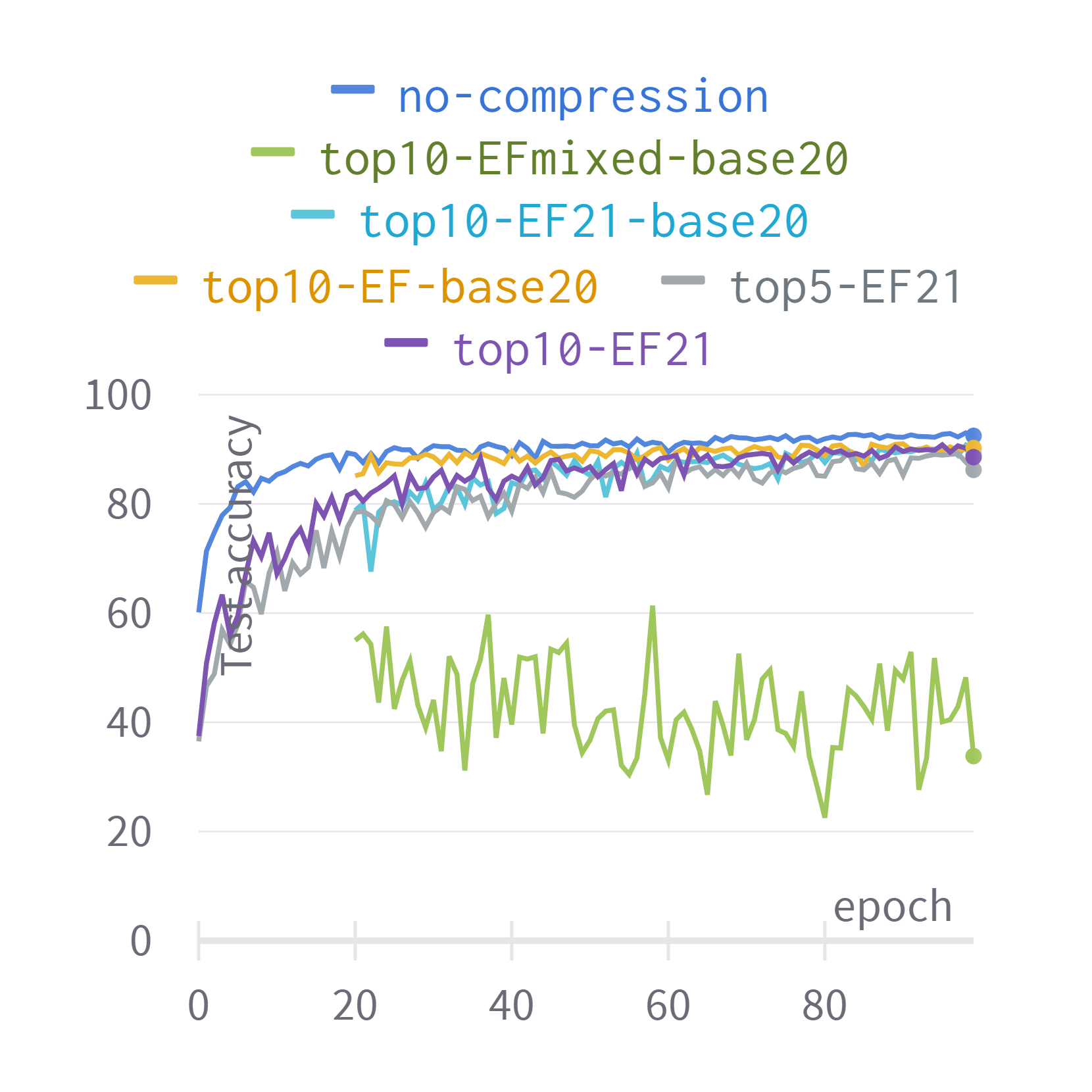}
         \caption{Test accuracy, \\compression off}
         \label{fig:ef-testacc-nocompression}
     \end{subfigure}%
     \hfill
     \begin{subfigure}{0.33\textwidth}
         \centering
         \includegraphics[width=\linewidth]{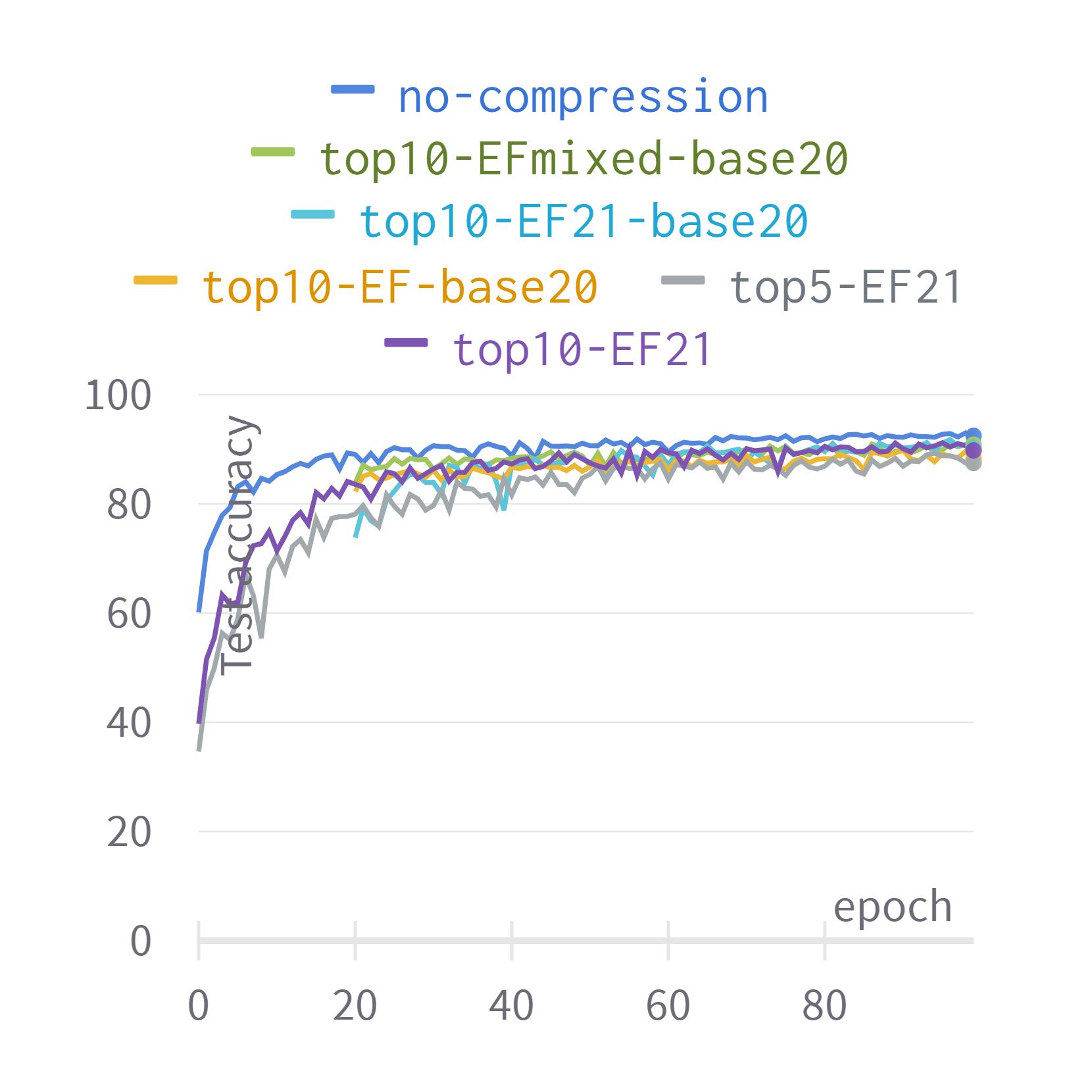}
         \caption{Test accuracy, \\with compression}
         \label{fig:ef-testacc-withcompression}
     \end{subfigure}%
    \caption[Error feedback experiments convergence on ResNet18 and CIFAR-10.]
    {Error feedback experiments convergence on ResNet18 and CIFAR-10.
    \\ Model-parallel degree is 4, with 3 compression operations used. Activations and gradients are compressed independently, with global EF batch buffer for each compression operator. Runs with suffix \textit{base20} use uncompressed baseline weights after 20 epochs.}
    \label{fig:results-ef}
\end{figure}

Table~\ref{table:results-ef} and Figure~\ref{fig:results-ef} shows training results for error feedback experiments. Results suggest that using EF in addition to TopK compression does not improve the model accuracy. Significant  difference of activations and gradients between batches may be why the EF-based approaches does not improve model quality when applied directly. Thus, adding errors to an example of a different target class may significantly affect the error in the buffer. This is probably why the global EF buffer fails to improve model performance in a model-parallel setup.

The unexpected yet positive finding is that the EF technique makes test accuracy results similar for uncompressed and compressed inference. Test accuracy without compression is only 1-2 percentage points below the one of compressed variant of EF or EF21, while for the plain TopK compression, the drop in test accuracy is at least 7-15 percentage points. We hypothesize that EF buffer is considered as noise to the model, and the model ignores such noise to the activations, becoming more robust and leading to comparable uncompressed and compressed test results.

\begin{table}[!h]
    \caption[AQ-SGD and TopK Experiments Training Results on ResNet18 and CIFAR-10]
    {AQ-SGD and TopK Experiments Training Results \\on ResNet18 and CIFAR-10}
    \label{table:results-resnet-aqsgd}
    \begin{center}
    \begin{tabularx}{0.85\textwidth}{X c c}
        \hline
        \makecell[l]{Compression\\Mode} & \makecell[l]{Test accuracy(\%), \\compression off} & \makecell[l]{Test accuracy(\%)\\with compression}\\
         \hline
         No compression & \textbf{93.00}  & \textbf{93.00} \\
         AQ-SGD + Top 50\%, warmup 10 & \textbf{92.44}    & \textbf{92.54}     \\
         AQ-SGD + Top 30\%, warmup 10 & \textbf{91.86}    & \textbf{91.61}    \\
         AQ-SGD + Top 20\%, warmup 10 & 90.82    & 87.8     \\
         AQ-SGD + Top 10\%, warmup 10 & 85.91    & 84.16     \\
         \hline
    \end{tabularx}
    \end{center}
    \begin{center}
    \small
    Each cell is a single run for 100 epochs.\\Warmup 10 runs use uncompressed baseline weights after 10 epochs.\\Model-parallel degree is 4, with 3 compression operations used.\\Activations and gradients are compressed independently, with AQ-SGD per-example buffer applied only for activations.
    \end{center}
\end{table}

\subsubsection{AQ-SGD with TopK compression}
\label{subsec:discussion_AQSGD}
Finally, Table~\ref{table:results-resnet-aqsgd} presents results of experiments based on AQ-SGD~\cite{wang2022fineAQSGD} approach with TopK compression used instead of quantization. 

\begin{figure}[htbp]
\captionsetup[subfigure]{justification=centering}
     \begin{subfigure}{0.33\textwidth}
         \centering
         \includegraphics[width=\linewidth]{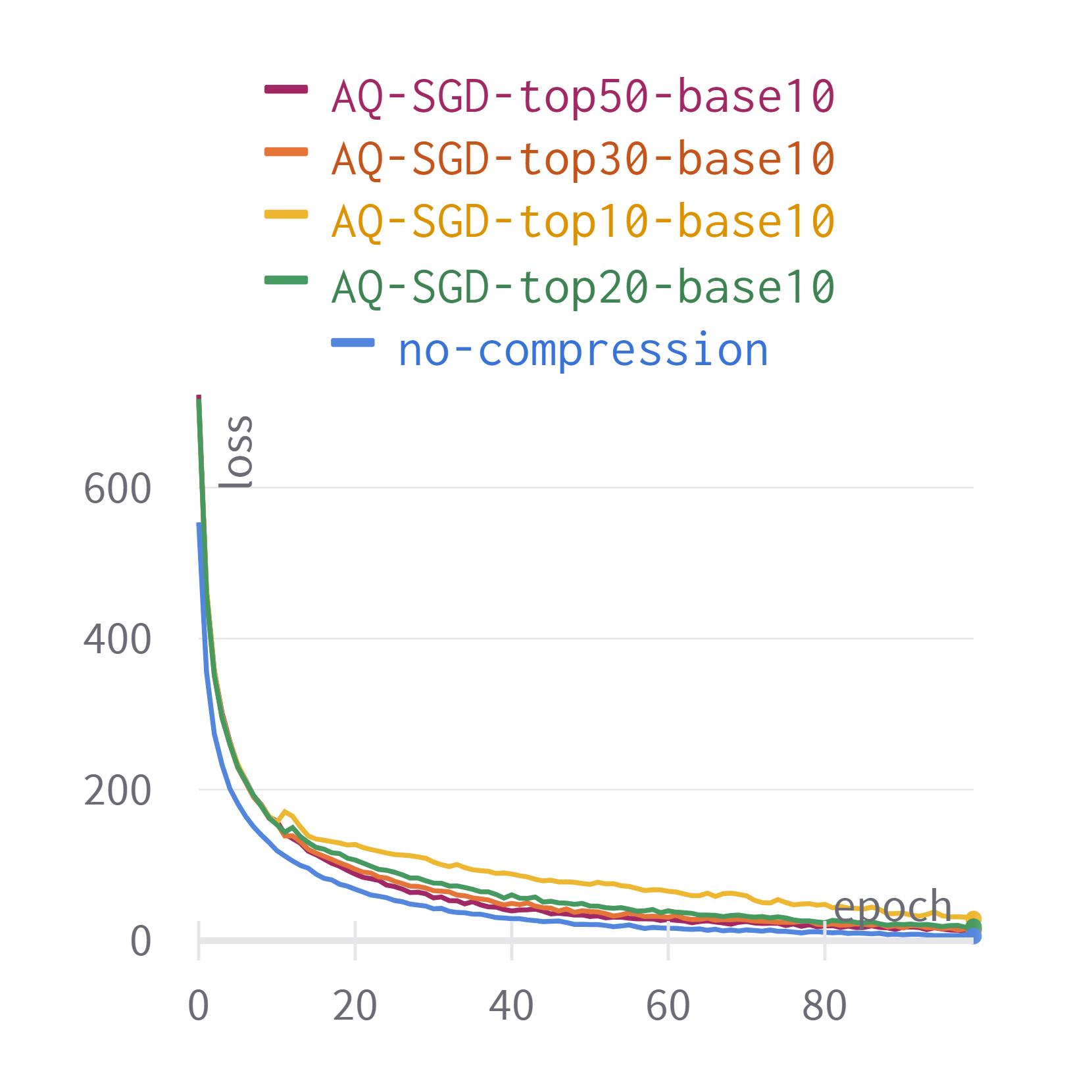}
         \caption{Train loss\\\phantom{..}}
         \label{fig:aqsgd-trainloss}
     \end{subfigure}%
     \hfill
     \begin{subfigure}{0.33\textwidth}
         \centering
         \includegraphics[width=\linewidth]{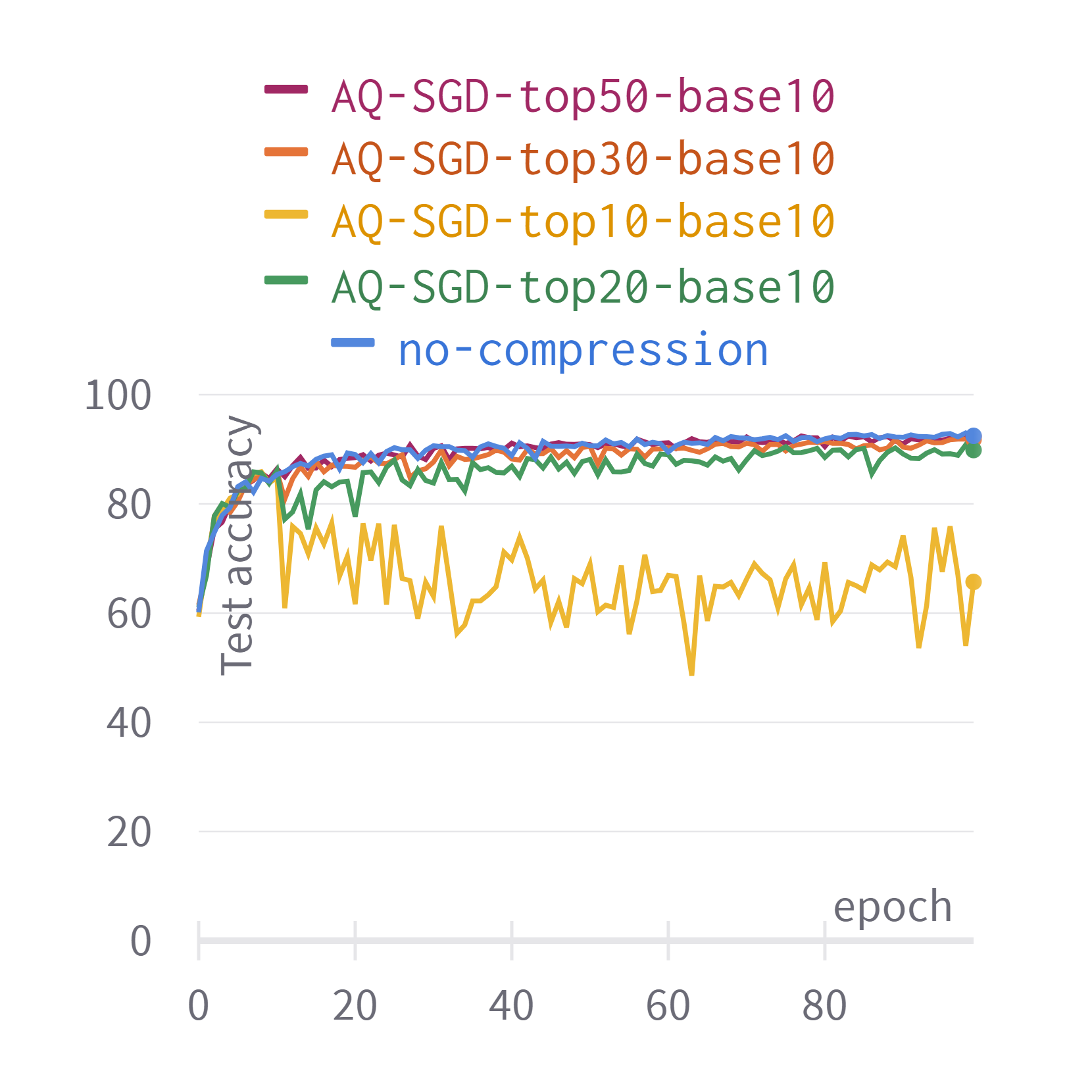}
         \caption{Test accuracy, \\compression off}
         \label{fig:aqsgd-testacc-nocompression}
     \end{subfigure}%
     \hfill
     \begin{subfigure}{0.33\textwidth}
         \centering
         \includegraphics[width=\linewidth]{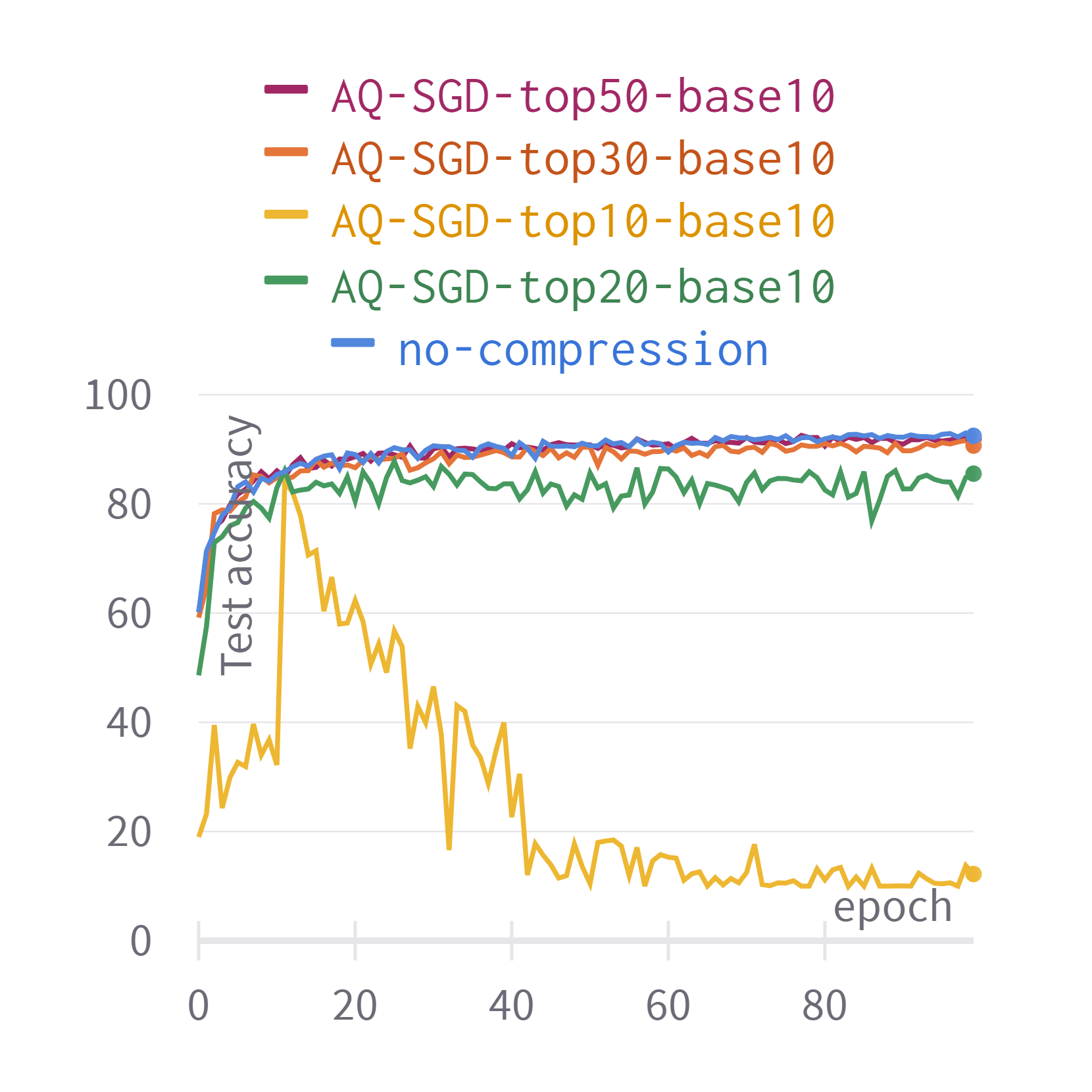}
         \caption{Test accuracy, \\with compression}
         \label{fig:aqsgd-testacc-withcompression}
     \end{subfigure}%
    \caption[AQ-SGD and TopK experiments convergence on ResNet18 and CIFAR-10.]
    {AQ-SGD and TopK experiments convergence on ResNet18 and CIFAR-10. Model-parallel degree is 4, with 3 compression operations used. Activations and gradients are compressed independently, with AQ-SGD per-example buffer applied only for activations. Runs with suffix \textit{base10} use uncompressed baseline weights after 10 epochs.}
    \label{fig:results-resnet-aqsgd}
\end{figure}

Learning curves in Figure~\ref{fig:results-resnet-aqsgd} suggest that convergence of the AQ-SGD method with TopK compression is not improved compared to the plain TopK compression. Test accuracy growth does not match the uncompressed baseline for Top10\% compression. We suppose that convergence for TopK compression used with the AQ-SGD approach is slow due to the biasedness of TopK compression.

\subsection{Fine-tuning GPT-2 on Wikitext}
\label{sec:gpt2-wikitext}

We fine-tuned pretrained GPT-2-small~\cite{radford2019languageGPT2} on Wikitext text dataset~\cite{wikitext103merity2016pointer}, wikitext-2-raw-v1 version. Quality was measured by test loss and perplexity metrics.
We fine-tuned the model for 4 epochs with a batch size of 8. We used Hugging Face model fine-tuning code\footnote{\url{https://github.com/huggingface/transformers/blob/main/examples/pytorch/language-modeling/run\_clm.py}} and integrated compression into it.
Experiments were conducted with Tesla P100-PCIE-16GB GPU. Model parallelism degree was set to four.

\begin{table}[!h]
    \caption[TopK Compression Fine-tuning Experiments Results on GPT-2 and Wikitext]
    {TopK Compression Fine-tuning Experiments Results \\on GPT-2 and Wikitext}
    \label{table:results-gpt2-topk}
    \begin{center}
    \begin{tabularx}{0.7\textwidth}{X c c}
        \hline
        \makecell[l]{Compression Mode} & \makecell[l]{Eval loss} & \makecell[l]{Perplexity}\\
         \hline
         No compression & \textbf{3.05}  & \textbf{21.01} \\
         Top 50\% & \textbf{3.11}   & \textbf{22.38}   \\
         Top 30\%  & \textbf{3.27}  & \textbf{26.28}     \\
         Top 20\% & 3.51   & 33.32    \\
         Top 10\% & 4.31    & 74.51     \\
         Top 10\% separate& 8.0   & 2990.16  \\
         \hline
    \end{tabularx}
    \end{center}
    \begin{center}
    \small
    Model-parallel degree is 4, with 3 compression operations used. TopK compression reuses TopK indices from activations to compress gradients.\\TopK separate mode compresses activations and gradients independently.
    \end{center}
\end{table}

\begin{figure}[!t]
\captionsetup[subfigure]{justification=centering}
     \begin{subfigure}{0.5\textwidth}
         \centering
         \includegraphics[width=\linewidth]{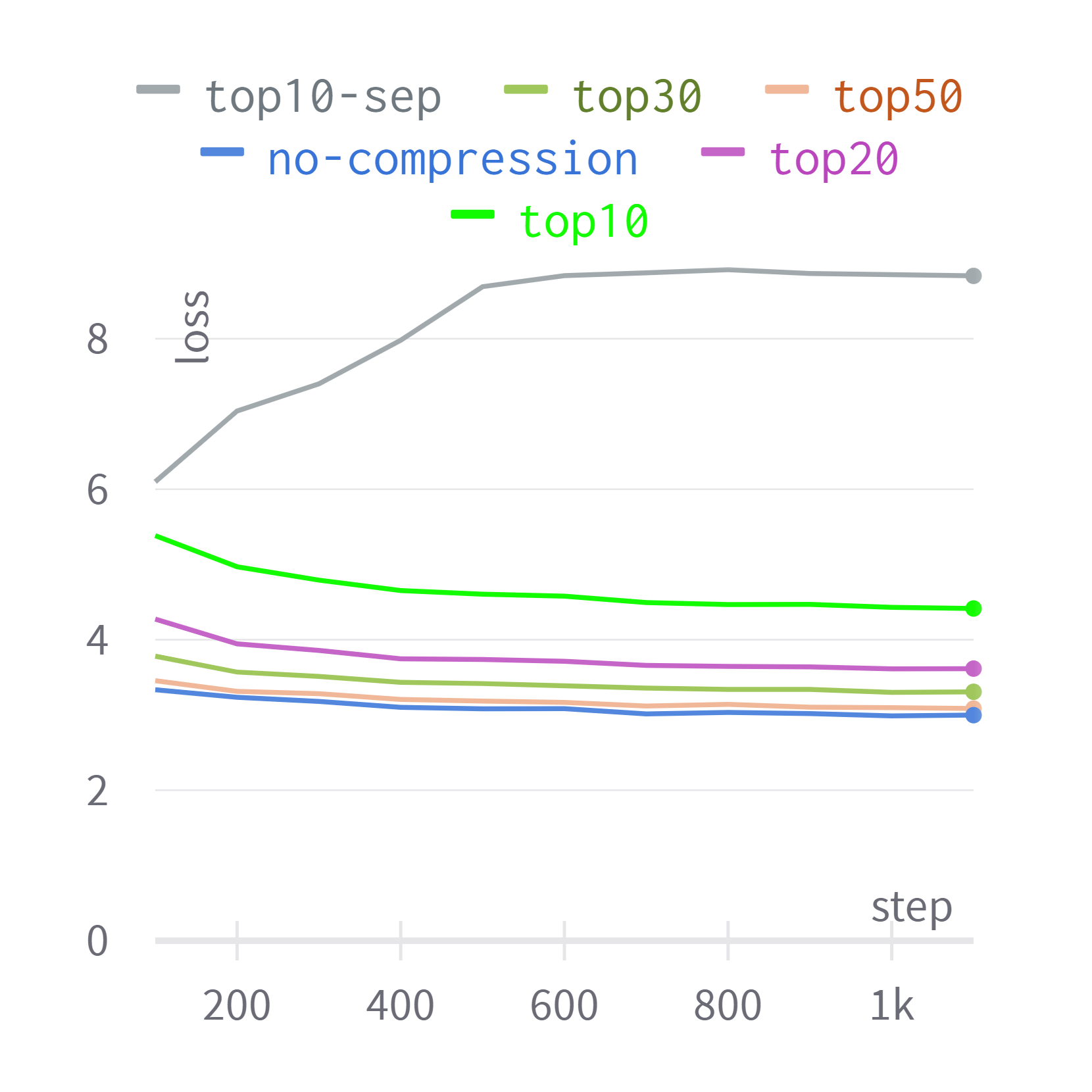}
         \caption{Train loss}
         \label{fig:gpt2-topk-trainloss}
     \end{subfigure}%
     \hfill
     \begin{subfigure}{0.5\textwidth}
         \centering
         \includegraphics[width=\linewidth]{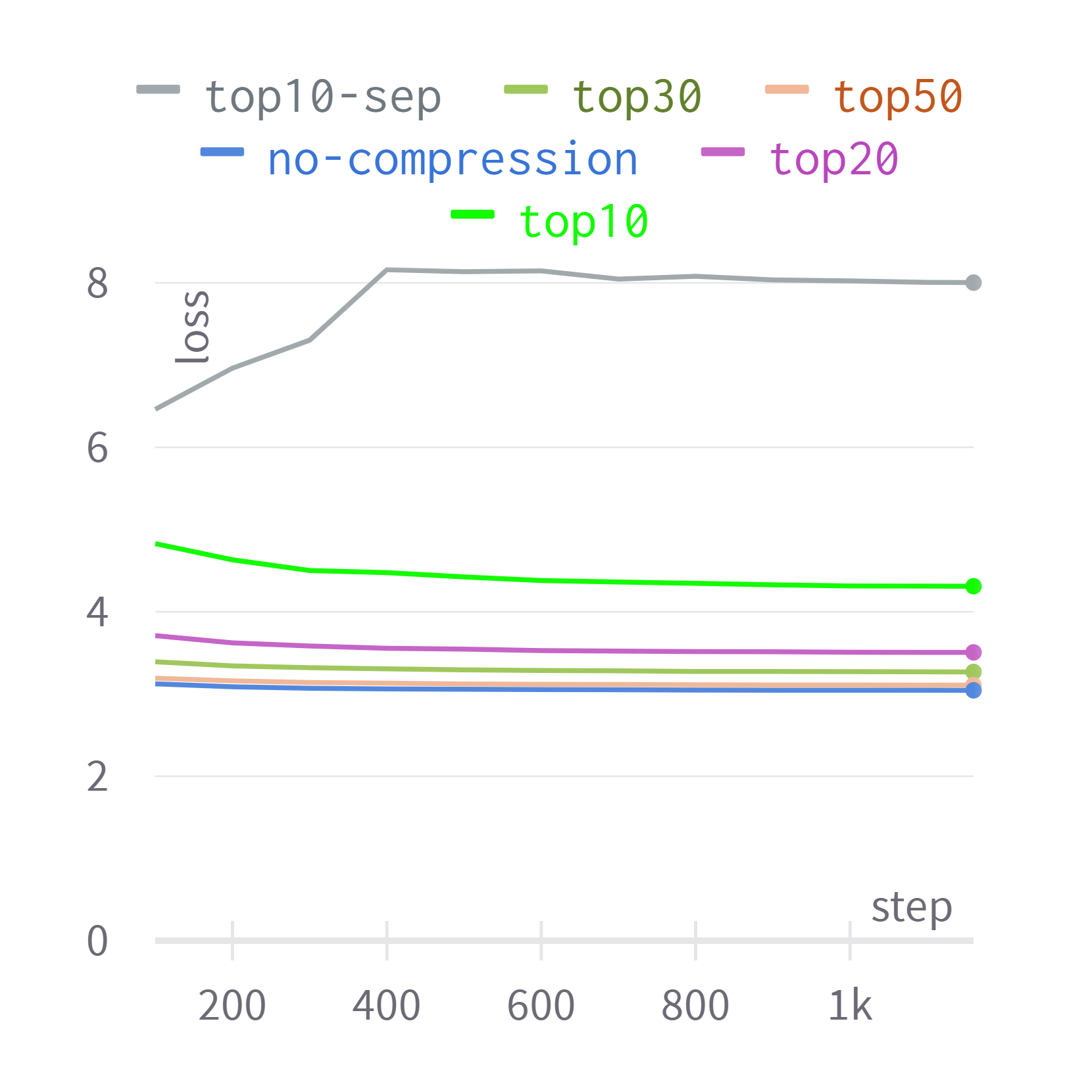}
         \caption{Eval loss}
         \label{fig:gpt2-topk-evalloss}
     \end{subfigure}%
    \caption[TopK experiments convergence on GPT-2 and Wikitext.]
    {TopK experiments convergence on GPT-2 and Wikitext.}
    \label{fig:results-gpt2-topk}
\end{figure}

Figure~\ref{fig:results-gpt2-topk} shows train and eval loss during fine-tuning. Table~\ref{table:results-gpt2-topk} presents fine-tuning results with TopK compression applied. The first observation is that TopK compression starts to increase validation loss significantly already from Top20\% compression. Compared to our experiments with ResNet18, where Top10\% compression resulted in satisfactory results, current setup does not tolerate such compression. Two factors may explain this behaviour: the architectural differences between transformer and CNN models, and the use of fine-tuning instead of training from scratch, which may hinder efficient learning of TopK compression distributions.

The second observation is regarding the independence of activations and gradients compression. In the fine-tuning scenario, compressing activations and gradients independently leads to very large eval loss and even divergence of the model. We hypothesize that the fact we use pretrained GPT-2 defines such behavior. For the pre-trained model, zeroing out a large ratio of activations affects gradients more than the actual training sample model is fine-tuned on. In our opinion, to prevent such behavior, reusing TopK\% indices is necessary to achieve good results.


\section{Related Works}
\label{section:literature_review}

This section describes related works on compression in distributed training. In general, compression techniques are used to reduce communication overhead, and can be applied to layers activations or their respective gradients.

\subsection{Gradients compression}
Gradients compression is naturally applied in the data-parallel approach. Several methods have been developed to reduce communication overhead without significantly decreasing model quality.

The most common way to reduce the space taken by floating point numbers is the quantization approach. Several works on robust quantization show that under certain assumptions and techniques used, quantized learning leads to the same convergence and model quality~\cite{seide20141,bernstein2018signsgd}. Alistarh et al.~\cite{alistarh2017qsgd} developed a framework for evaluating convergence of quantization-based gradient compression methods. Quantization methods with distribution-aware choice of quantization levels achieve even better practical results applied to large models training~\cite{han2015deepquantizationaware,hong2022daquantizationaware,fu2020dontwastebits}. Quantization methods have the advantage of being unbiased, hence being able to converge as with uncompressed training.

Another class of compression methods is based on low-rank approximation of gradients vector. Wang et al.~\cite{wang2018atomolowrank} propose compression by sparsification of singular values of gradients. PowerSGD by Vogels et al.~\cite{vogels2019powersgd} compresses gradients even more with the concept of power iteration. Such methods also apply to model parallelism: Optimus-CC framework shows superior training time results with PowerSGD-based gradients compression \cite{song2023optimus}. However, the decomposition of gradients requires additional computation time for compression.

To cope with errors introduced by gradient compression, error feedback (EF) techniques have been proposed to improve convergence. Initially proposed by Seide et al.~\cite{seide20141} as a heuristic, EF has shown to be an effective method for improving convergence for data-parallel tasks. The newer EF21 approach improves EF by communicating changes of gradients \cite{richtarik2021ef21}. MARINA~\cite{gorbunov2021marina} and DIANA~\cite{mishchenko2019distributedDIANA} use compression of gradient differences to lower convergence bounds for strongly convex and non-convex objectives in data-parallel setup.
Optimus-CC~\cite{song2023optimus} also compresses differences of gradients between mini-batches, resembling the EF21 approach, but applied to model parallelism. Overall, error feedback techniques achieve better theoretical and practical results in data parallel setting, but only a few works use EF with model parallelism.

Finally, various sparsification methods are also used for gradient compression.  
Beznosikov et al. ~\cite{beznosikov2020biased} analyze the theoretical bounds of biased methods. They also propose improved TopK compression with exponential dithering to achieve more economy in communication with comparable model convergence in a data-parallel setup. However, sparsification (for example, TopK compression) requires to send not only the selected data, but also the indices, which increases communication cost.

\subsection{Activations compression}
Similarly to gradient compression, activation compression could also be applied in both data- and model-parallel setups. Usually, it is used to reduce memory footprint on inference, as well as communication time in the case of model parallelism.

As the sizes of activations can be significant, some works compress activations even without model parallelism. Compressed activations are used to compute model gradients during a backward pass. AC-GC~\cite{evans2021acgc} uses fixpoint quantization to compress activations up to 15 times with only 0.1\% average accuracy loss. Fu et al.~\cite{fu2020dontwastebits} also use distribution-aware quantization to compress activations and gradients, decreasing memory consumption up to 2-4 times. A novel approach LLM-int8()~\cite{dettmers2022llm} uses 8-bit quantization for speeding up inference of large language models without quality degradation. This approach also considers outliers values passed in the original fp32 format.

A notable application of the error feedback approach to activations compression has been introduced in AQ-SGD model-parallel compression framework~\cite{wang2022fineAQSGD}. This method communicates quantized differences of activations between pipeline blocks, achieving up to 8.5 times increase of total training throughput over slow networks.

Experimental work on activations compression methods by Bian et al.~\cite{bian2023doesactivationcompression} compares popular compressors for model-parallel training and draws several conclusions on the practical application of activations compression. In particular, they observe that learning-based autoencoder compression performs best in terms of convergence and training throughput compared to quantization and TopK compression.


\section{Conclusion}
\label{section:conclusion}

In this work, we conducted experiments on communication compression of activations and gradients for model-parallel neural networks distributed training. In particular, we explored how quantization, TopK compression, and error feedback-based approaches perform in several machine learning tasks.

We observed that model gradients are more sensitive to compression than activations as shown in quantization experiments. We also empirically showed that TopK compression for both activation and gradients can be applied with at most Top 10\%  to achieve comparable model quality.
We evaluated error feedback techniques and noticed no significant improvement in convergence with TopK compression of activations and gradients in the model-parallel setup. However, error feedback techniques helped to overcome inference performance drop without compression applied. Finally, we reported that the AQ-SGD~\cite{wang2022fineAQSGD} approach could not be applied with TopK compression stronger than $K=30\%$ to achieve satisfactory convergence.

Several potential limitations of our research should be acknowledged. Firstly, since model parallelism is usually applied for large models, this work lacks experiments with training state-of-the-art large models.
Secondly, we test only one model parallelism pipeline and hyperparameters configuration for each experiment, which should be extended in the future work.

We propose several research directions based on our work: experiments with large language models, reducing AQ-SGD approach memory footprint, and explore more biased compression techniques apart from TopK.


\section*{Acknowledgements}

We deeply thank the most irreplaceable and the most secretive colleague from Yandex.Research for insightful discussions during our research. We also thank Anton Antonov for providing computational resources on early stages of the project.

\bibliographystyle{splncs04}
\bibliography{ref}

\end{document}